\documentclass[10pt,twocolumn,letterpaper]{article}

\usepackage[pagenumbers]{cvpr} %

\usepackage[dvipsnames]{xcolor}

\usepackage{xr}

\definecolor{cvprblue}{rgb}{0.21,0.49,0.74}
\usepackage[pagebackref,breaklinks,colorlinks,citecolor=cvprblue]{hyperref}

\usepackage{glossaries}
 
\newacronym{ai}{AI}{Artificial Intelligence}
\newacronym{dl}{DL}{Deep Learning}
\newacronym{dnn}{DNN}{Deep Neural Network}
\newacronym{ood}{OOD}{Out Of Distribution}
\newacronym{lrp}{LRP}{Layer-wise Relevance Propagation}
\newacronym{xai}{XAI}{eXplainable Artificial Intelligence}
\newacronym{crp}{CRP}{Concept Relevance Propagation}
\newacronym{crv}{CRV}{Concept Relevance Vector}
\newacronym{cav}{CAV}{Concept Activation Vector}
\newacronym{pcrv}{PCRV}{Prototypical Concept Relevance Vector}
\newacronym{ml}{ML}{Machine Learning}
\newacronym{gmm}{GMM}{Gaussian Mixture Model}
\newacronym{auc}{AUC}{Area Under the Curve}
\newacronym{se}{SE}{Standard Error}
\newacronym{cnn}{CNN}{Convolutional Neural Network}
\newacronym{llm}{LLM}{Large Language Model}
\newacronym{dora}{DORA}{Data-agnOstic Representation Analysis}
\newacronym{invert}{INVERT}{Inverse Recognition}
\newacronym{ea}{EA}{Extreme-Activation}
\newacronym{ours}{PURE}{\underline{Pu}rifying \underline{Re}presentations}

\def\paperID{*****} %
\def\confName{CVPR}
\def\confYear{2024}

\title{PURE: Turning Polysemantic Neurons Into Pure Features by Identifying Relevant Circuits}

\author{
Maximilian Dreyer$^{1}$,
Erblina Purelku$^1$,
Johanna Vielhaben$^{1}$,\\
Wojciech Samek$^{1,2,3,\dagger}$,
Sebastian Lapuschkin$^{1,\dagger}$\\
$^1$ Fraunhofer Heinrich Hertz Institute,
$^2$ Technical University of Berlin, \\
$^3$ BIFOLD – Berlin Institute for the Foundations of Learning and Data\\
$^\dagger${\small corresponding authors:} 
{\tt\small  \{wojciech.samek\,|\,sebastian.lapuschkin\}@hhi.fraunhofer.de}
}

\begin{document}
\maketitle

\begin{abstract}
The field of mechanistic interpretability aims to study the role of individual neurons in Deep Neural Networks.
Single neurons, however, 
have the capability to act polysemantically and encode for multiple (unrelated) features, which renders their interpretation difficult.
We present a method for disentangling polysemanticity of any Deep Neural Network
by decomposing a polysemantic neuron into multiple monosemantic ``virtual'' neurons.
This is achieved by identifying the relevant sub-graph (``circuit'') for each ``pure'' feature.
We demonstrate how our
approach allows us to find and disentangle various polysemantic units of ResNet models trained on ImageNet.
While evaluating feature visualizations using CLIP, 
our method effectively disentangles representations, improving upon methods based on neuron activations.
Our code is available at \url{https://github.com/maxdreyer/PURE}.

\end{abstract}

\section{Introduction}
The field of \gls{xai} aims to increase the transparency of \glspl{dnn}.
Several \gls{xai} works study the role of a model's latent neurons and their interactions \cite{olah2017feature,olah2020zoom},
which recently developed into the sub-field of \emph{mechanistic interpretability}.
Neurons are commonly viewed as feature extractors corresponding to human-interpretable concepts \cite{olah2017feature,achtibat2023attribution,bau2017network}. 
However,
neurons can be \emph{polysemantic},
meaning that they extract multiple (unrelated) features,
which adds ambiguity to their interpretability.
Other \gls{xai} works study \emph{circuits}, \ie, distinct sub-graphs of a network performing specific sub-tasks, which recently became popular for \glspl{llm} \cite{elhage2021mathematical,wang2023interpretability,hanna2024does}. 
Notably, the interpretability of a circuit depends on the interpretability of its units, which again can be polysemantic.
In applications, such as knowledge discovery or validation of \glspl{dnn} in safety-critical tasks,
high latent interpretability is crucial for \gls{xai} usefulness~\cite{bykov2023dora,dreyer2023understanding}.
\begin{figure}
    \centering
    \includegraphics[width=0.95\linewidth]{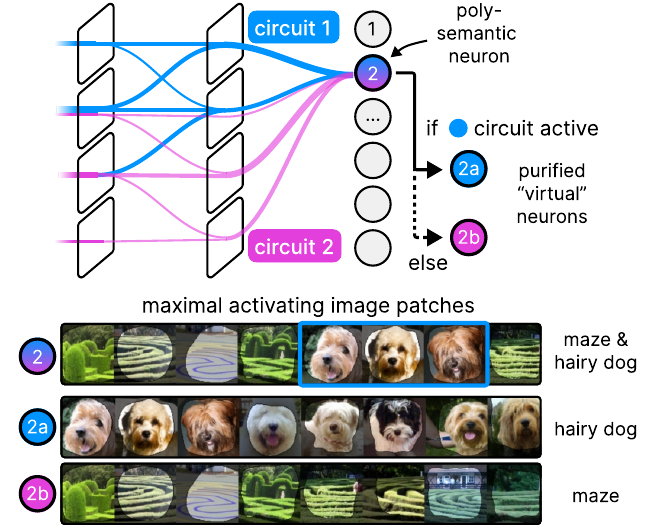}
    \caption{Distinct circuits exist for each feature of a polysemantic neuron. With PURE, we propose to split a polysemantic neuron into multiple pure ``virtual'' ones, one for each circuit. Here, we disentangle the maximally activating sample (patches) of neuron 2 into its two pure features: ``hairy dog''  (2a) and ``maze'' (2b).}
    \label{fig:intro}
\end{figure}
In this work,
we build on the assumption that for each monosemantic (``pure'') feature a unique sub-graph exists.
Identifying the active circuits then allows disentangling a polysemantic unit into multiple ``virtual'' pure units (with one circuit each),
as shown in \cref{fig:intro} where we disentangle a neuron encoding for dog and maze features.

To that end,
we introduce \gls{ours}, 
a post-hoc approach for increasing interpretability of latent representations by
disentangling polysemantic neurons into pure features.
\gls{ours} is based on discovering the relevant (active) circuit of each semantics,
which is identified via a partial backward pass.
Through the means of foundational models,
we measure a significant increase in interpretability of ResNet~\cite{he2016deep} models after applying \gls{ours},
also improving upon activation-based approaches.

\begin{figure*}
    \centering
    \includegraphics[width=1.0\linewidth]{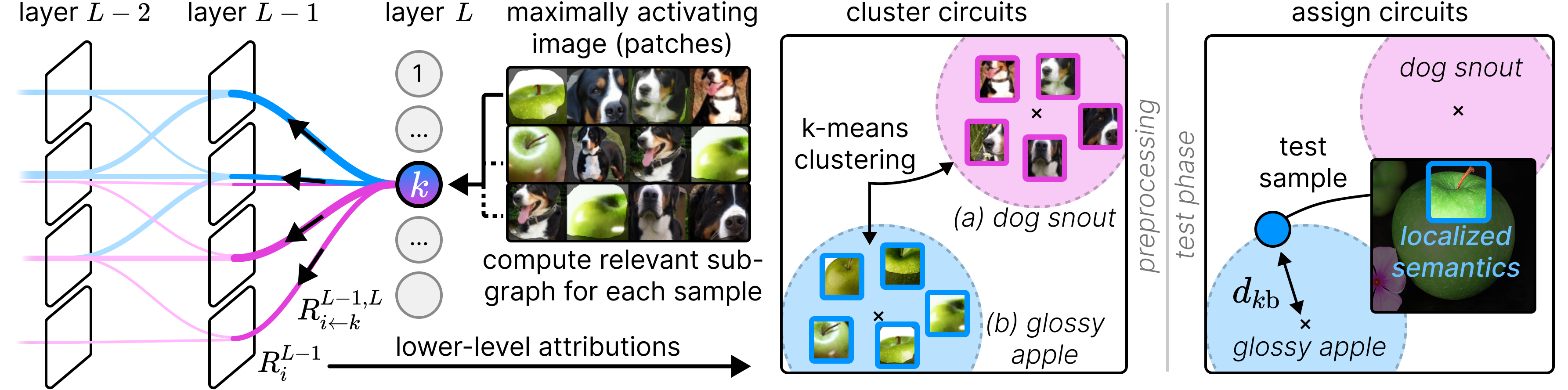}
    \caption{\gls{ours} detects circuits using lower-level neuron attributions for the $n_\text{ref}$ most activating input samples in a preprocessing step. 
    For polysemantic neurons,
    we assume distinct active circuits for each semantics,
    which are found through clustering attributions with $k$-means.
    During test phase, the active circuit can be assigned post-hoc for any new test sample by identifying the closest circuit.}
    \label{fig:method}
\end{figure*}

\section{Related Work}
Various works have shown
that neurons in \glspl{dnn} encode for distinct features that can often be interpreted by humans~\cite{bau2017network,achtibat2023attribution,bykov2024labeling}.
However,
besides redundancies in representations,
an occurring problem is the polysemanticity of neurons.
As such, 
interpretation of the latent space is difficult.
For concept-based explanations, feature visualizations are confusing, or ambiguous as it might be unclear,
which semantics are actually present.
Further,
semantics can be overlooked, \eg, when one feature is more dominant than another~\cite{nguyen2016multifaceted}.
As a way out,
some works propose to find more meaningful directions~\cite{kim2018interpretability,fel2024holistic} or subspaces in latent space~\cite{vielhaben2023multi},
but they often require pre-defined concepts or reconstruct the latent space only partially.
To resolve poly-semanticity on a neuron level,
O‘Mahony \etal \cite{O'Mahony_2023_CVPR} propose to find directions (\ie, a linear combination of neurons) based on latent activations.
Instead of activations (that depend on \emph{all} present input features),
\gls{ours} is based on neuron-specific circuits which is more specific to the role of a neuron and leads to an improved disentanglement.

Circuits, in general, are viewed as the sub-graphs of a neural network architecture \cite{wang2023interpretability} that perform a specific task.
They further consist of a set of linked features and the weights between them \cite{olah2020zoom}. 
In recent work, circuit analysis \cite{Raukur2022TowardTA} has been extended beyond convolution-based architectures~\cite{cammarata2020thread:} to, \eg, transformer-based models \cite{elhage2021mathematical,wang2023interpretability,hanna2024does}.
The discovery of circuits has been partially automatized both for computer vision~\cite{rajaram2023automatic} and language models~\cite{conmy2024towards}.

\section{Method}

For \gls{ours},
we view circuits as directed acyclic graphs 
that consist of nodes $R_j^{l}$ for neuron $j$ in layer $l$ and edges $R_{i\leftarrow j}^{l-1, l}$ connecting nodes $j$ and $i$ of adjacent layers.
We hypothesize that polysemanticity of a neuron arises because multiple circuits share one node, as illustrated in \cref{fig:method}. 
\gls{ours} disentangles these circuits from a neuron perspective by clustering neuron $k$'s functional connectivity with neurons of lower layers.
This process involves two steps: (1) computing circuits, and (2) replacing a shared neuron with multiple virtual neurons for each circuit through clustering.

\begin{figure*}[t]
    \centering
    \includegraphics[width=0.94\linewidth]{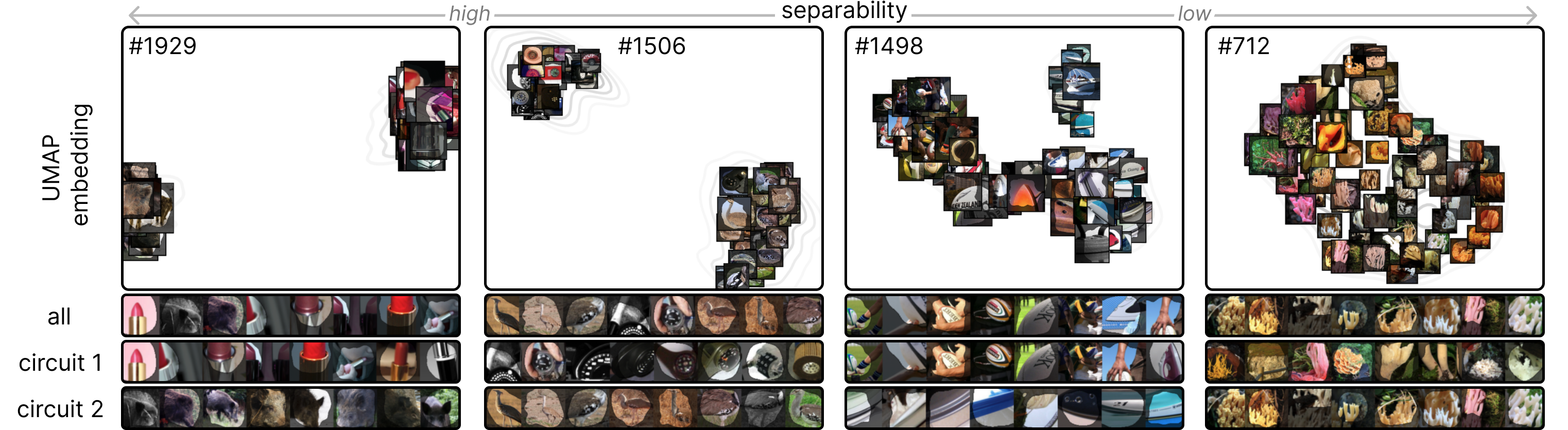}
    \caption{Applying \gls{ours} to neurons with varying degree of polysemanticity: We show UMAP embeddings with the maximally activating image patches, and the resulting reference sets before and after purification when identifying two circuits via $k$-means.}
    \label{fig:qualitative}
\end{figure*}
\textbf{Step 1) Computing Circuits:}
To compute the edges of a circuit,
we \emph{explain} the activation of a neuron 
and attribute lower-level neurons.
This naturally fits the idea of backpropagation-based feature attribution methods, specifically LRP~\cite{bach2015pixel}.
LRP allows to efficiently backpropagate attribution scores through the network layer by layer, beginning at a latent neuron
until the desired (input) layer is reached~\cite{bach2015pixel}. 
Concretely,
the relevance $R^{l}_j$ of an upper layer neuron $j$ is generally distributed to lower-level neurons $i$ as

\begin{equation}
    R^{l - 1, l}_{i\leftarrow j} = \frac{z_{i\rightarrow j}^{l - 1, l}}{z_j^{l}} R^{l}_j 
\end{equation}
with $z_{i\rightarrow j}^{l - 1, l}$ contributing to $z_j^{l} = \sum_i z_{i\rightarrow j}^{l-1,l}$ in the forward pass.
Note that
multiple refined ways to define %
$R^{l - 1, l}_{i\leftarrow j}$
are proposed in literature for different layer types ~\cite{montavon2019layer}.
These relevance ``messages'' $R^{l - 1, l}_{i\leftarrow j}$ refer to the \emph{edges} of a circuit.

The circuit \emph{nodes} are then characterized by node attributions computed via aggregation of the relevance messages:
\begin{equation}
\label{eq:node_relevance} 
    R^{l-1}_i = \sum_j R^{l - 1, l}_{i\leftarrow j},
\end{equation}
which reduces to $R^{L-1}_i = R^{L - 1, L}_{i\leftarrow k}$ for the first backpropagation step when explaining neuron $k$ in layer $L$ as in \cref{fig:method}.

For simplicity, we investigate in the following
only nodes of the next lower-level layer $L-1$.
It is to note,
that other methods besides LRP can be used for attributions here~\cite{fel2024holistic}.
We therefore default to Gradient$\times$Activation as an efficient and universal attribution method implementing a simple LRP variant in ReLU-\glspl{dnn}~\cite{shrikumar2017learning}.
Thus,
the circuit computation for neuron $k$ in layer $L$ simplifies for \gls{ours} to
 
\begin{equation}
\label{eq:node_relevance_simple}
    R^{L-1}_i =  A_i^{L-1} \frac{\partial A_k^L}{\partial A_i^{L-1}},
\end{equation}
for circuit nodes $i$ in lower-level layer $L-1$, with activation $A_k^L$ of neuron $k$ in layer $L$, and partial derivative $\partial/\partial A_i^{L-1}$ \wrt lower-level layer activations $A_i^{L-1}$.

\textbf{Step 2) Assigning Circuits:}
We represent a neuron by its most activating input samples.
Then, for each of the top-$n_\text{ref}$ activating input samples of a polysemantic neuron $k$ in layer $L$,
we compute the lower-level attributions $R^{L-1}_j$ for all $n$ neurons of layer $L-1$  as given by \cref{eq:node_relevance_simple} in a partial backward pass.
If neuron $j$ uses different circuits among the reference samples, we expect to see distinct clusters in the attribution vectors  $\mathbf{R}^{L-1}
\in\mathbb{R}^n$.
To validate this assumption, we visualize a 2D UMAP~\cite{mcinnes2018umap} embedding of $\mathbf{R}^{L-1}$
 in \cref{fig:method}.
To find the distinct clusters, we use 
$k$-means clustering which results in centroids representing new virtual neurons for each circuit.
For a new test sample,
we can then identify the active circuit by computing $\mathbf{R}^{L-1}$ and assigning it to the closest cluster centroid, \ie, virtual neuron.

\section{Experiments}

    We address the following research questions:
    \begin{enumerate}
        \item \textbf{(Q1)}~Can we find and purify polysemantic neurons? 
        \item \textbf{(Q2)}~How effective is \gls{ours} in disentangling representations compared to other approaches?
    \end{enumerate}

    \paragraph{Experimental Setting}
    We investigate the neurons in the penultimate layer of ResNet-34/50/101 models~\cite{he2016deep} pre-trained~\cite{wightman2021resnet} on the ImageNet~\cite{russakovsky2015imagenet} dataset,
    and evaluate interpretability using the foundational models of CLIP~\cite{radford2021learning} and DINOv2~\cite{oquab2023dinov2}.
    We perform the analysis on the $n_\text{ref} = 100$ maximally activating input samples (based on max-pooling) for each neuron on the ImageNet test set.
    To generate feature visualizations, we crop samples such that only the important part of a neuron's semantics remains~\cite{achtibat2023attribution}, as illustrated in \cref{fig:method} (\emph{right}) and detailed in \cref{sec:appendix:feature_visualizations}.

\subsection{From Polysemanticity to Pure Features (Q1)}
\label{sec:exp:distribution}

    We begin with the quest to find polysemantic units by studying their feature visualizations, \ie, their most activating image patches.
    As a quantitative and objective measure for monosemanticity,
    we evaluate CLIP embeddings, 
    where visually similar feature visualizations presumably result in small embedding distances~\cite{khalibat2023identifying, zhang2018unreasonable}.
    Concretely,
    for each neuron $k$ we compute the distance matrix
    \begin{equation}
    \label{eq:distance_matrix}
        \mathbf{D}^k_{ij} = \sqrt{\left(\mathbf{e}_i^\text{CLIP} - \mathbf{e}_j^\text{CLIP} \right)^2}  
    \end{equation}
    between the CLIP embeddings $\mathbf{e}_{i}^\text{CLIP}$ of all pairs of feature visualizations (cropped reference samples) $i$ and $j$.

    To optimize the process of finding polysemantic units,
    we perform $k$-means clustering on the CLIP embeddings with a fixed number of two clusters.
    Then, 
    inter- and intra-cluster distances $\rho_k$ for neuron $k$ are computed as
    \begin{equation} 
    \label{eq:distances}
        \rho_k^\text{intra} = \frac{\sum_{i, j\neq i}^{n_\text{ref}} \mathbf{D}^k_{ij} \mathbf{1}_{c_i = c_j}}{\sum_{i, j\neq i}^{n_\text{ref}}\mathbf{1}_{c_i = c_j}}, \quad
        \rho_k^\text{inter} = \frac{\sum_{i, j}^{n_\text{ref}} \mathbf{D}^k_{ij} \mathbf{1}_{c_i \neq c_j}}{\sum_{i, j}^{n_\text{ref}}\mathbf{1}_{c_i \neq c_j}}
    \end{equation}
    with indicator function $\mathbf{1}_{c_i = c_j}$ equaling one if feature visualizations $i$ and $j$ have the same cluster index $c$, and zero else.
    A large difference $\rho_k^\text{inter} - \rho_k^\text{intra}$ indicates clearly separated clusters with different semantics.
    
    In \cref{fig:qualitative},
    neurons of varying degree of polysemanticity, as given by $\rho_k^\text{inter} - \rho_k^\text{intra}$, are shown for ResNet-50.
    Here, we also depict UMAP embeddings based on \gls{ours} attributions given by \cref{eq:node_relevance_simple},
    and the feature visualizations before and after applying \gls{ours}.
    Note,
    that for \gls{ours},
    we here disentangle one neuron into two virtual ones by clustering the lower-level attributions.
    As visible,
    polysemantic neurons such as \texttt{\#1929} and \texttt{\#1506} exist and can be effectively disentangled.
    The disentangled semantics can be visually different (``lipstick'' and ``boar'') or more related (``white spots'' on dark circular objects or bird wings).
    On the other hand,
    we can also find rather monosemantic neurons,
    \eg, \texttt{\#1498} and \texttt{\#712}, that encode for lines and coral structure, respectively.
    More examples and more detailed results \wrt the distribution of polysemanticity are given in \cref{sec:appendix:distribution}.

\subsection{Evaluating Feature Purification (Q2)}
    \label{sec:evaluation:monosemanticity}
    \gls{ours} aims to turn polysemantic neurons into purer virtual neurons that are easier to interpret,
    and works
    by computing and identifying the relevant circuits based on lower-level attributions $R^{L-1}_i$ as by \cref{eq:node_relevance_simple}.
    Alternatively,
    O‘Mahony \etal \cite{O'Mahony_2023_CVPR} propose activation-based disentanglement,
    where the activations $A^L_i$ resulting from reference samples are clustered.
    When applied to a (polysemantic) neuron,
    both methods should result in more meaningful reference sample subsets for each disentangled feature, as, \eg, in \cref{fig:qualitative}. 
    
    To systematically evaluate the interpretability of newly disentangled representations, 
    we compute inter- and intra-cluster distances as defined in \cref{eq:distances} using CLIP for the resulting sets of feature visualizations (cropped reference samples). 
    Ideally,
    intra-cluster distances are low and inter-cluster distances are high,
    indicating well separated feature visualizations and more meaningful representations.
    \begin{figure}[t]
        \centering
        \includegraphics[width=0.95\linewidth]{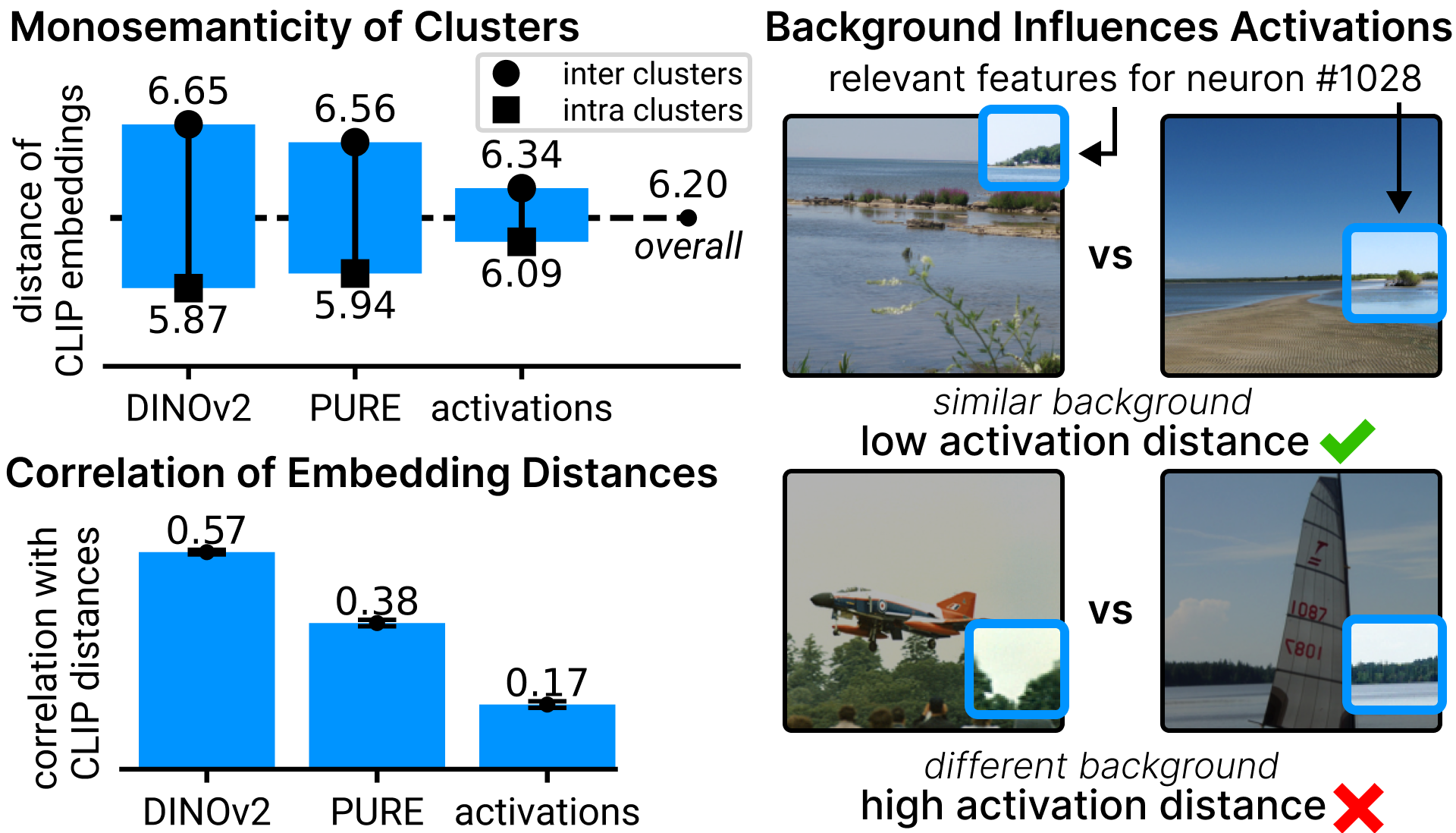}
        \caption{\gls{ours} leads to more interpretable representations as measured via CLIP embedding distances on feature visualizations (\emph{top left}),
        thereby improving upon activation-based clustering and reaching almost DINOv2 scores.
        (\emph{Bottom left}): Distances of CLIP embeddings between feature visualizations correlate with \gls{ours} embeddings significantly more than activations.
        (\emph{Right}): Activations tend to overestimate distances when unrelated features vary. 
        }
        \label{fig:quantitative}
    \end{figure}
    The resulting CLIP embedding distances are reported in \cref{fig:quantitative} (\emph{top left}) for \gls{ours} and activation-based disentanglement for ResNet-101.
    As another baseline,
    we perform clustering on DINOv2 embeddings for the cropped reference samples.
    Here, DINOv2 represents an ideal visual separation,
    which is, however, computationally expensive as it requires both the computation of the cropped reference samples and a DINOv2 forward pass.
    The results show,
    that \gls{ours} leads to more disentangled representations than activation-based clustering, and is performance-wise close to DINOv2.
    Note that so far, clusters are computed using $k$-means with $k=2$,
    but the same trends hold for different $k\in\{3, 4, 5\}$ and other ResNet architectures, as discussed in \cref{sec:appendix:evaluation}.

   In a second experiment,
   we dive deeper into why \gls{ours} attributions are more meaningful than latent activations.
   We thus investigate whether when two feature visualizations are similar according to CLIP,
   they are also similar according to \gls{ours} attributions or activations.
   Concretely,
   for feature visualization pairs,
   we compute CLIP embedding distances via \cref{eq:distance_matrix} and distances between \gls{ours} attributions as well as activations,
   and finally measure the correlation between the resulting distances of different methods.
    Please note, again,
    CLIP (and DINOv2) embeddings refer to the \emph{cropped} reference samples,
    whereas \gls{ours} and activations are computed on the \emph{full} reference samples.
    As shown in \cref{fig:quantitative} (\emph{bottom left}),
    \gls{ours} has higher alignment to CLIP compared to activations.
    We observe that activations lead to deviating distance scores in some cases,
    especially when the relevant semantics are very localized in reference samples, as shown in \cref{fig:quantitative} (\emph{right}) for neuron \texttt{\#1028} encoding for ``vegetation on horizon''.
    Notably,
    activations take into account \emph{all} present features in the \emph{full} reference sample,
    which influences distances when unrelated features (\eg, airplanes or boats) vary between samples.
    Whereas, conditional attributions as used by \gls{ours} are more specific to the actual task of a neuron.
    Correlation results for the other ResNet architectures and more examples when \gls{ours} or activation-based clustering diverges from CLIP are given in \cref{sec:appendix:evaluation}.

\section{Limitations and Future Work}
    So far,
    we assumed that embeddings of foundational models can be seen as ideal indicators for human interpretability and disentanglement.
    In future work,
    evaluation in controlled settings or using human feedback will be valuable.

    Regarding \gls{ours},
    a large ablation study will be interesting, \eg,
    for clustering on full circuits (instead of only lower-level layer attributions),
    and 
    using different feature attribution methods or other clustering approaches than $k$-means.
    Notably,
    \gls{ours} requires a partial backward pass to disentangle a neuron, which is computationally slightly more demanding than activation-based disentangling.
    
    It will be interesting to further study the advantages of purified units for \gls{xai} tools such as concept-based explanations, concept discovery and probing, or model correction.

\section{Conclusion}
    We introduce \gls{ours},
    a novel method for turning polysemantic neurons into multiple purer ``virtual'' neurons by identifying the active characteristic circuit of each pure feature. 
    The purification of latent units allows to better understand latent representations,
    which is especially interesting for the growing and promising field of concept-based \gls{xai}.
    Using foundational models for evaluation,
    \gls{ours} results in significantly more purified features than activation-based approaches,
    which are less neuron-specific.
    We believe that our work will raise interest in investigating the benefits of cleaner representations for, \eg, concept discovery, concept probing or model correction.

    \subsection*{Acknowledgements}
    This work was supported by
    the Federal Ministry of Education and Research (BMBF) as grant BIFOLD (01IS18025A, 01IS180371I);
    the German Research Foundation (DFG) as research unit DeSBi (KI-FOR 5363);
    the European Union’s Horizon Europe research and innovation programme (EU Horizon Europe) as grant TEMA (101093003);
    the European Union’s Horizon 2020 research and innovation programme (EU Horizon 2020) as grant iToBoS (965221);
    and the state of Berlin within the innovation support programme ProFIT (IBB) as grant BerDiBa (10174498).

{
    \small
    \bibliographystyle{ieeenat_fullname}
    \bibliography{main}
}
\cleardoublepage
\appendix
\renewcommand\thefigure{A.\arabic{figure}}    
\renewcommand\thetable{A.\arabic{table}}
\setcounter{figure}{0}
\setcounter{table}{0}
\renewcommand\theequation{A.\arabic{equation}}
\setcounter{equation}{0}
    \section{Appendix}
    In the appendix,
    we provide additional results for the main manuscript.
    Concretely,
    we give details on how we compute feature visualizations in \cref{sec:appendix:feature_visualizations}.
    Secondly, in \cref{sec:appendix:distribution}, the distribution of polysemanticity throughout neurons of ResNet~\cite{he2016deep} models is shown in higher detail.
    Further,
    we provide additional examples of resulting disentangled representations when applying \gls{ours} in  \cref{sec:appendix:pure_examples}.
    Lastly,
    \cref{sec:appendix:evaluation} provides more results for evaluating feature interpretability after purification of neurons.

\subsection{Feature Visualizations}
\label{sec:appendix:feature_visualizations}

An important part in understanding neurons are feature visualizations that aim to communicate the underlying semantics or concept of a neuron.
In literature,
either real data samples, or synthetically generated samples are used for visualization purposes~\cite{olah2017feature}.
Throughout our experiments, 
we utilize reference samples from the original dataset to render visualizations that are as natural as possible, ideally staying in-distribution \wrt the foundational models of CLIP~\cite{radford2021learning} and DINO~\cite{oquab2023dinov2} in evaluations.

\begin{figure}[t]
        \centering
        \includegraphics[width=1.0\linewidth]{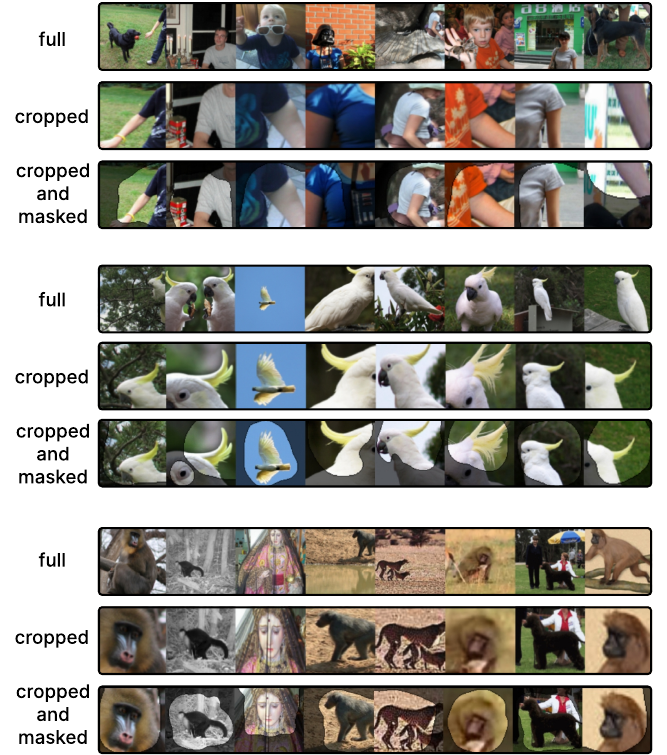}
        \caption{Feature visualizations for neurons \texttt{\#1028} (\emph{top}), \texttt{\#1029} (\emph{middle}) and \texttt{\#1030} (\emph{bottom}) for full, cropped-only and cropped as well as masked reference samples.
        It is visible that cropping improves visualizations by removing irrelevant and distracting image parts not relevant for a neuron semantics.
        }
        \label{fig:appendix:feature_vis}
        \end{figure}

When using reference samples from the dataset, it is crucial to crop images to the actually important part for a neuron, as semantics can be very localized,
as, \eg, visible in \cref{fig:appendix:feature_vis} when comparing ``full'' against ``cropped'' samples.
In order to detect this ``relevant'' part,
Achtibat \etal~\cite{achtibat2023attribution} propose to explain neuron activations using LRP~\cite{bach2015pixel} in a first step, which results in neuron-specific input feature attributions.
Specifically for convolutional layers,
the maximum activation of a feature map is explained.
In a second step,
the attributions are smoothed using a Gaussian filter with kernel size $K$, normalized to a maximum value of one, and the image cropped and masked to include only attributions above a threshold of $T$.
For the masked part, black color is overlaid with 40\,\% transparency.

When evaluating visualizations with CLIP and DINO,
we use cropped samples with $K=5$ and $T=0.01$.
We refrain from masking samples, as masks could introduce out-of-distribution data.
For visualizations in plots,
we include masks as proposed by \cite{achtibat2023attribution} using $K=51$ and $T=0.01$.

\clearpage
\clearpage
\subsection{Distribution of Polysemanticity}
\label{sec:appendix:distribution}
    In \cref{sec:exp:distribution},
    we measure the degree of monosemanticity by using CLIP embeddings and computing distances in embedding space for feature visualization pairs of a neuron.
    Small distances between embeddings presumably correspond to visually similar feature visualizations.
    
    For polysemantic neurons, where, \eg, two monosemantic features superimpose, we would see two distinct clusters for each set of reference samples (of each pure feature) in a UMAP~\cite{mcinnes2018umap} embedding. 
    The inter-cluster distance, in this case, will be high, whereas intra-cluster distance will be low.
    In the following,
    we apply $k$-means clustering ($k=2$) on CLIP embeddings for all neurons \wrt the 100 most activating reference samples, and measure the overall as well as inter- and intra-cluster distances as given by \cref{eq:distances}.
    
\begin{figure*}[t]
        \centering
        \includegraphics[width=1.0\linewidth]{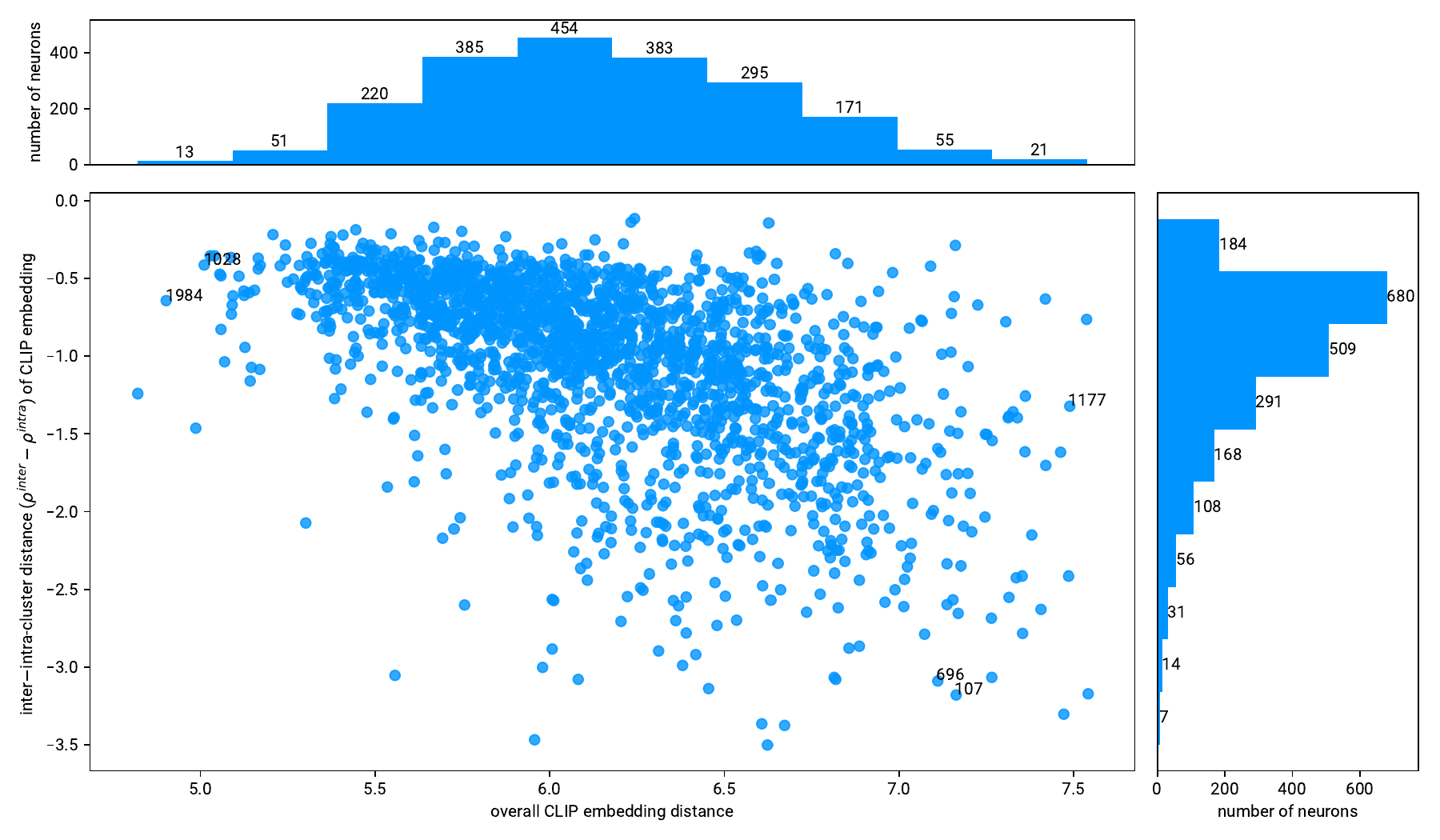}
        \includegraphics[width=1.0\linewidth]{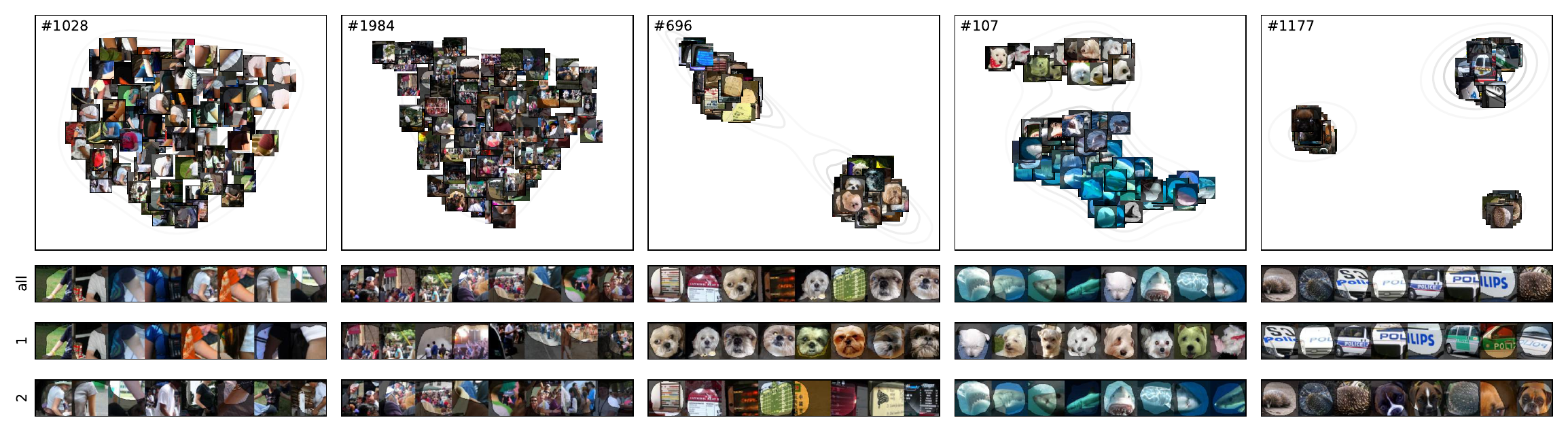}
        \caption{Distribution of CLIP embeddings distances for all neurons of the ResNet-50 model. We show the overall distances between feature visualizations of a neuron on the horizontal axis, and the difference between inter- and intra-cluster distance after clustering visualizations into two clusters on the vertical axis.
        (\emph{Bottom}): Examples are given for neurons with various degrees of polysemanticity. UMAP embeddings for \gls{ours} attributions as well as reference samples for the original and two virtual neurons are shown when applying \gls{ours}.
        }
        \label{fig:appendix:distribution50}
    \end{figure*}

    In \cref{fig:appendix:distribution50},
    we show the distribution of distances (intra- and inter-cluster distance difference, and overall distance) for ResNet-50.
    It is apparent,
    that for most neurons, clustering improves the visual similarity of feature visualizations according to CLIP.
    However,
    for some (a few hundred of the 2048 neurons), a larger improvement can be seen,
    indicating more polysemantic neurons.

     Examples with UMAP embeddings and clustered reference samples for neurons of varying degrees of polysemanticity are shown in \cref{fig:appendix:distribution50} (\emph{bottom}).
    Neurons \texttt{\#1028} and \texttt{\#1984} have low overall CLIP embedding distance and correspond to monosemantic features, \eg, ``human arms'' and ``human crowds'', respectively.
    Further, neurons \texttt{\#696} and \texttt{\#107} have large improvement in embedding distance when clustered,
    indicating polysemanticity, \eg, ``dog face'' vs. ``text/horizontal lines'' and ``shark under water'', respectively.
    Lastly,
    we show neuron \texttt{\#1177} with high overall distances, where clustering does not strongly decrease distances.
    This is apparently due to three existing semantics in the neuron, where clustering using two clusters is not optimal for disentanglement.

\begin{figure*}[t]
        \centering
        \includegraphics[width=1.0\linewidth]{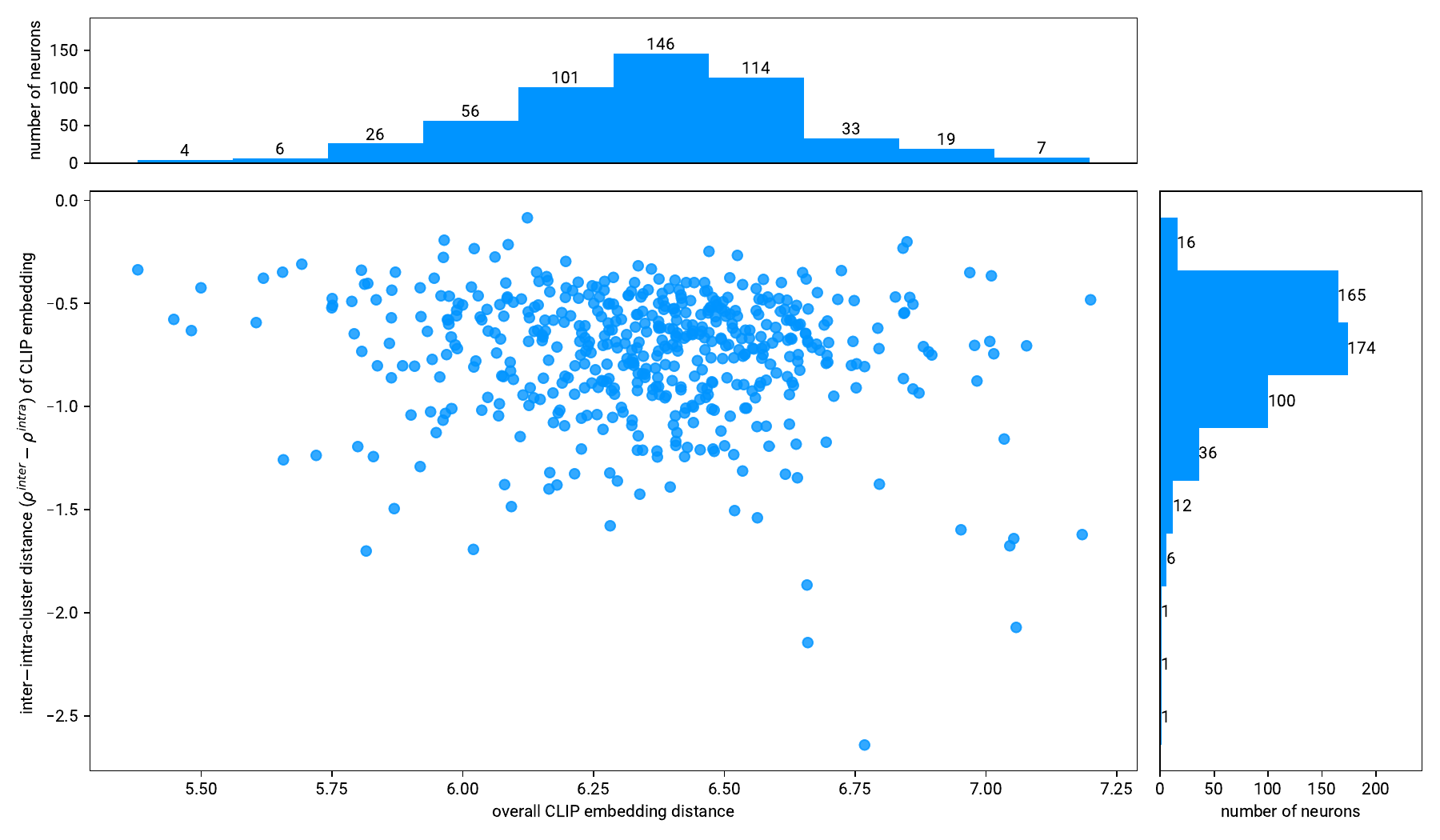}
                \includegraphics[width=1.0\linewidth]{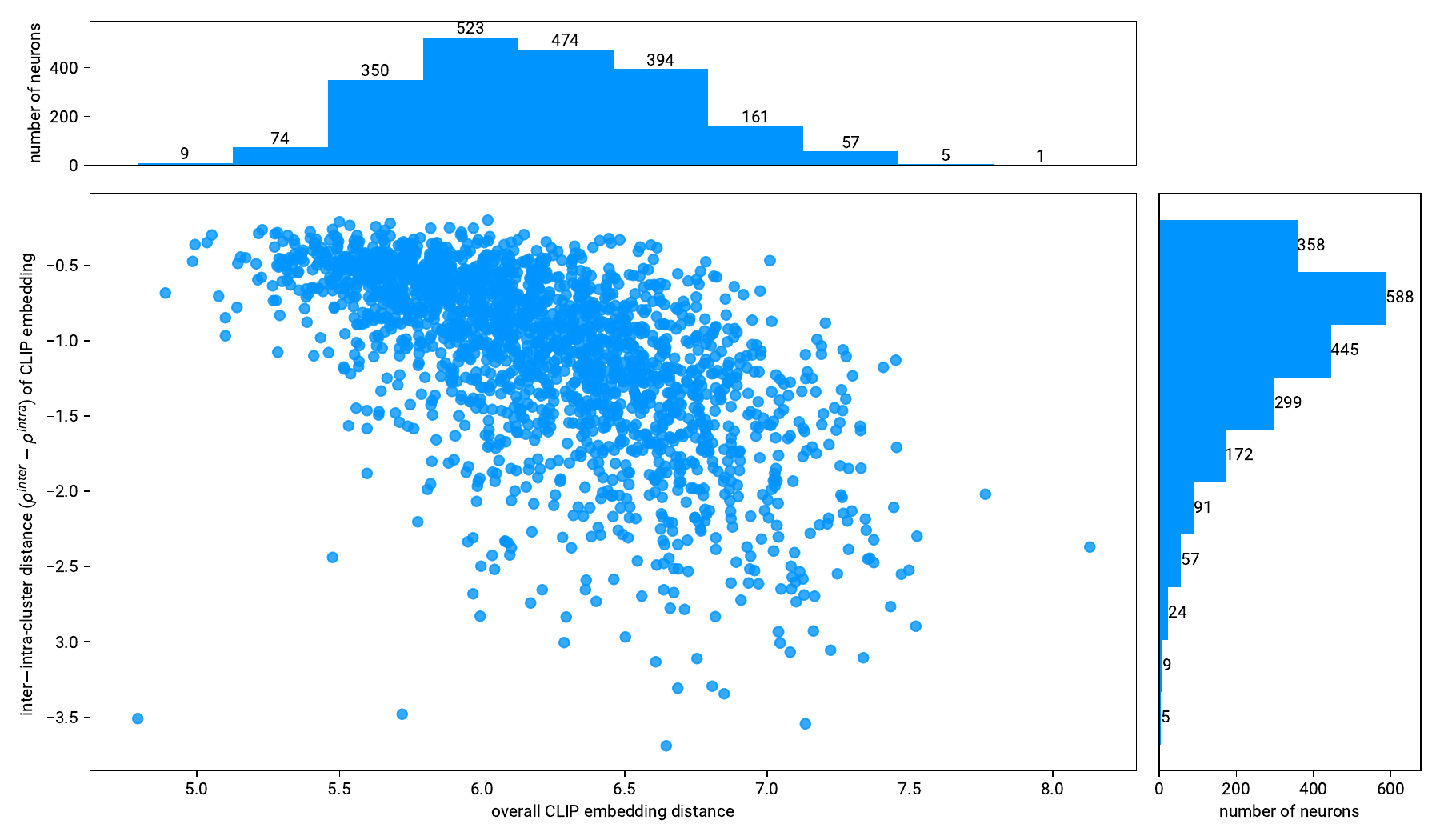}
        \caption{Distribution of CLIP embeddings distances for all neurons of the ResNet-34 (\emph{top}) and ResNet-101 (\emph{bottom}) model. We show the overall distances between feature visualizations of a neuron on the horizontal axis, and the difference between inter- and intra-cluster distance after clustering visualizations into two clusters on the vertical axis.}
        \label{fig:appendix:distribution34}
    \end{figure*}

    Additionally,
    we provide distribution plots for ResNet-34 and ResNet-101 in \cref{fig:appendix:distribution34}.
    It is to note,
    that ResNet-34 only consists of 512 instead of 2048 neurons in the penultimate layer.
    Comparing ResNet-34 and ResNet-100 distributions,
    improvements in CLIP embedding distances through clusters seem to be lower for ResNet-34, potentially indicating a smaller degree of polysemanticity.
    However,
    we leave comparison across architectures for future work.

    \subsection{Examples for Applying PURE}
    \label{sec:appendix:pure_examples}
    In the following,
    we present additional examples when applying \gls{ours} to neurons with different separability levels, which, for neurons with high separability score, leads to multiple virtual (ideally more disentangled) neurons.
    
    As in \cref{sec:exp:distribution},
    we create two virtual neurons using $k$-means for each neuron of a ResNet-50 in the penultimate layer. We then visualize the UMAP embeddings of the \gls{ours} attributions given by \cref{eq:node_relevance_simple} and the resulting clustered feature visualizations.
    
    \begin{figure*}[t]
            \centering
        \includegraphics[width=1.0\linewidth]{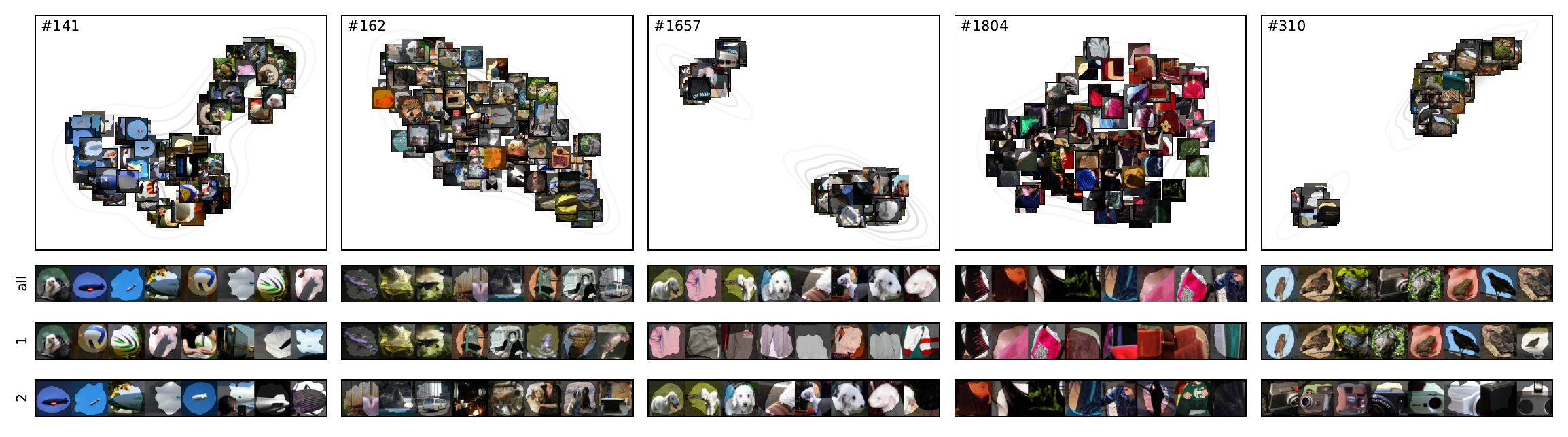}
            \caption{Examples of applying \gls{ours} to \emph{randomly chosen} neurons. Here we see the UMAP embeddings of the maximally activating patches, and the resulting reference sets before and after applying purification when identifying two circuits via $k$-means. In ResNet-50 neurons with different level of polysemanticity can be found.}
        \label{fig:appendix:pure_examples:random}
    \end{figure*}

        \begin{figure*}[t]
            \centering
        \includegraphics[width=1.0\linewidth]{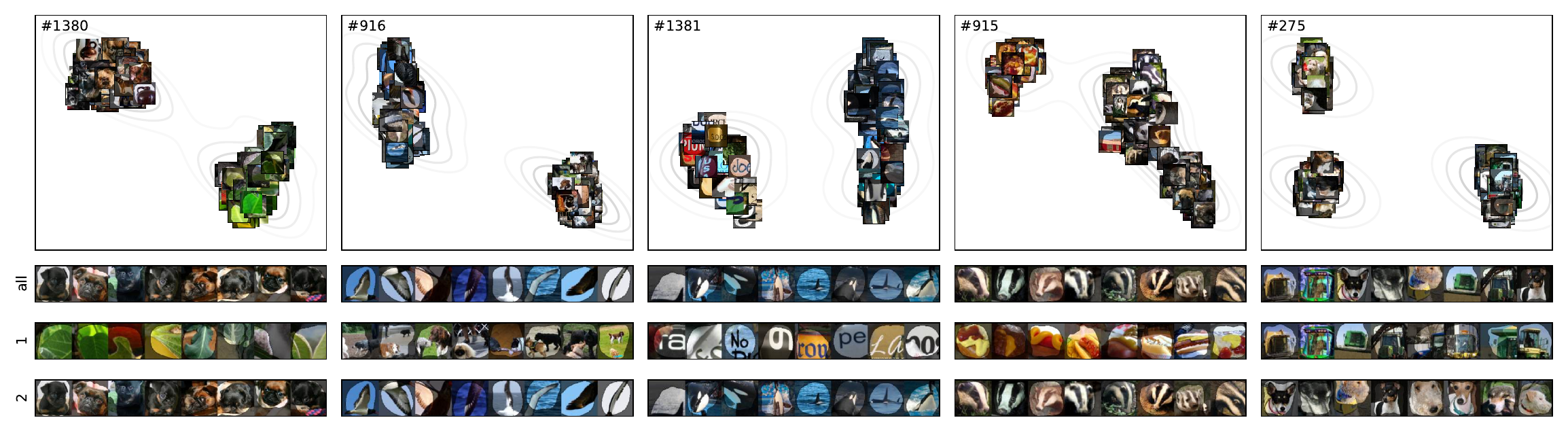}
            \caption{Examples of applying \gls{ours} to neurons with \emph{high degree of polysemanticity}. Here we see the UMAP embeddings of the maximally activating patches, and the resulting reference sets before and after applying purification when identifying two circuits via $k$-means. We show that neurons with high degree of polysemanticity can be successfuly disentangled into two (or more) monosemantic neurons.}
        \label{fig:appendix:pure_examples:separable}
     \end{figure*}
     
        \begin{figure*}[t]
            \centering
        \includegraphics[width=1.0\linewidth]{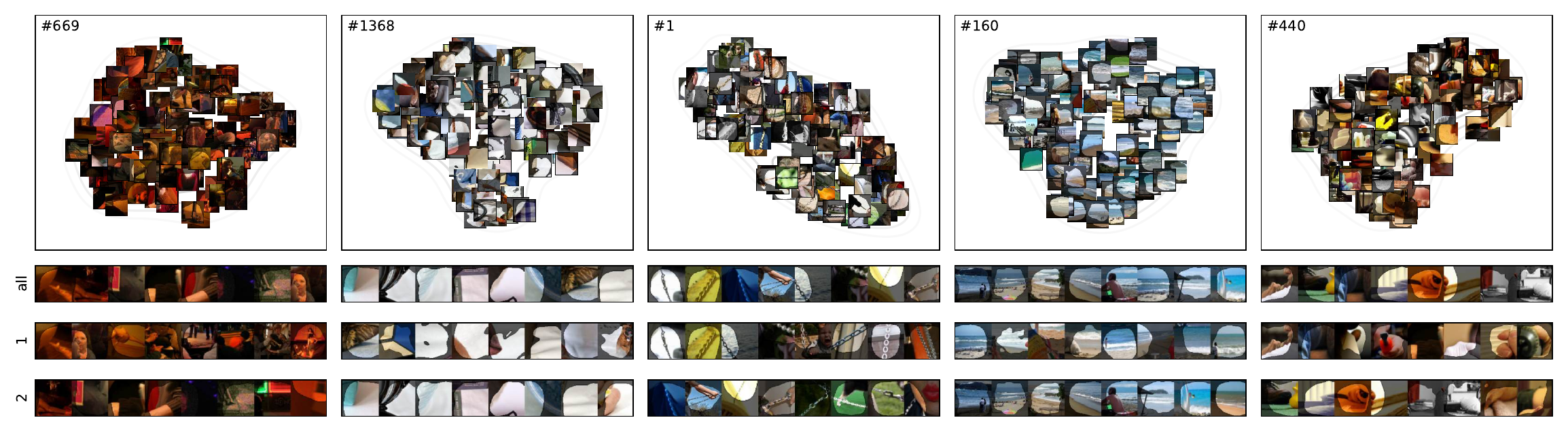}
            \caption{Examples applying \gls{ours} to neurons with \emph{low degree of polysemanticity}. Here we see the UMAP embeddings of the maximally activating patches, and the resulting reference sets before and after applying purification when identifying two circuits via $k$-means. The neurons encode one or similar features in this case, resulting in one circuit to be identified.}
        \label{fig:appendix:pure_examples:nonseparable}
    \end{figure*}
    In \cref{fig:appendix:pure_examples:random} we show five \emph{randomly sampled} neurons which exhibit different levels of separability; we see that both monosemantic (\texttt{\#162} and \texttt{\#1804}) and polysemantic (\texttt{\#141}, \texttt{\#1657}, and \texttt{\#310}) neurons can be found.
    Regarding the polysemantic units \texttt{\#1657} and \texttt{\#310},
    \gls{ours} leads to well separated reference samples.
    For \texttt{\#141},
    three semantics seem to exist, where clustering with two clusters does not optimally disentangle the unit.

    In \cref{fig:appendix:pure_examples:separable}, we focus on neurons with higher degree of polysemanticity (indicated by a large inter-cluster distance and low intra-cluster distance on \gls{ours} embeddings), which can be meaningfully disentangled into multiple virtual monosemantic neurons using \gls{ours}. 
    For instance, \eg neuron \texttt{\#1381} encodes both for ``printed letters'' and ``orca'', which \gls{ours} effectively disentangled.
    Similarly, neurons \texttt{\#916} and \texttt{\#915} refers to semantics such as ``two dogs'' and ``bird wings'' and of ``badger'' and ``ketchup and mustard in hot dog'' features, respectively.
    
    \cref{fig:appendix:pure_examples:nonseparable} illustrates examples of neurons encoding for pure features with small differences in inter- and intra-cluster distances (with a low level of separability). Noteworthy, instances include neurons such as \texttt{\#1}, which represents ``chain'', or neuron \texttt{\#160}, encoding ``seashore/shore''.

\clearpage
\clearpage
    \subsection{Evaluating Neuron Purification using CLIP}
    \label{sec:appendix:evaluation}
    
\begin{figure*}[t]
        \centering
        \includegraphics[width=1.0\linewidth]{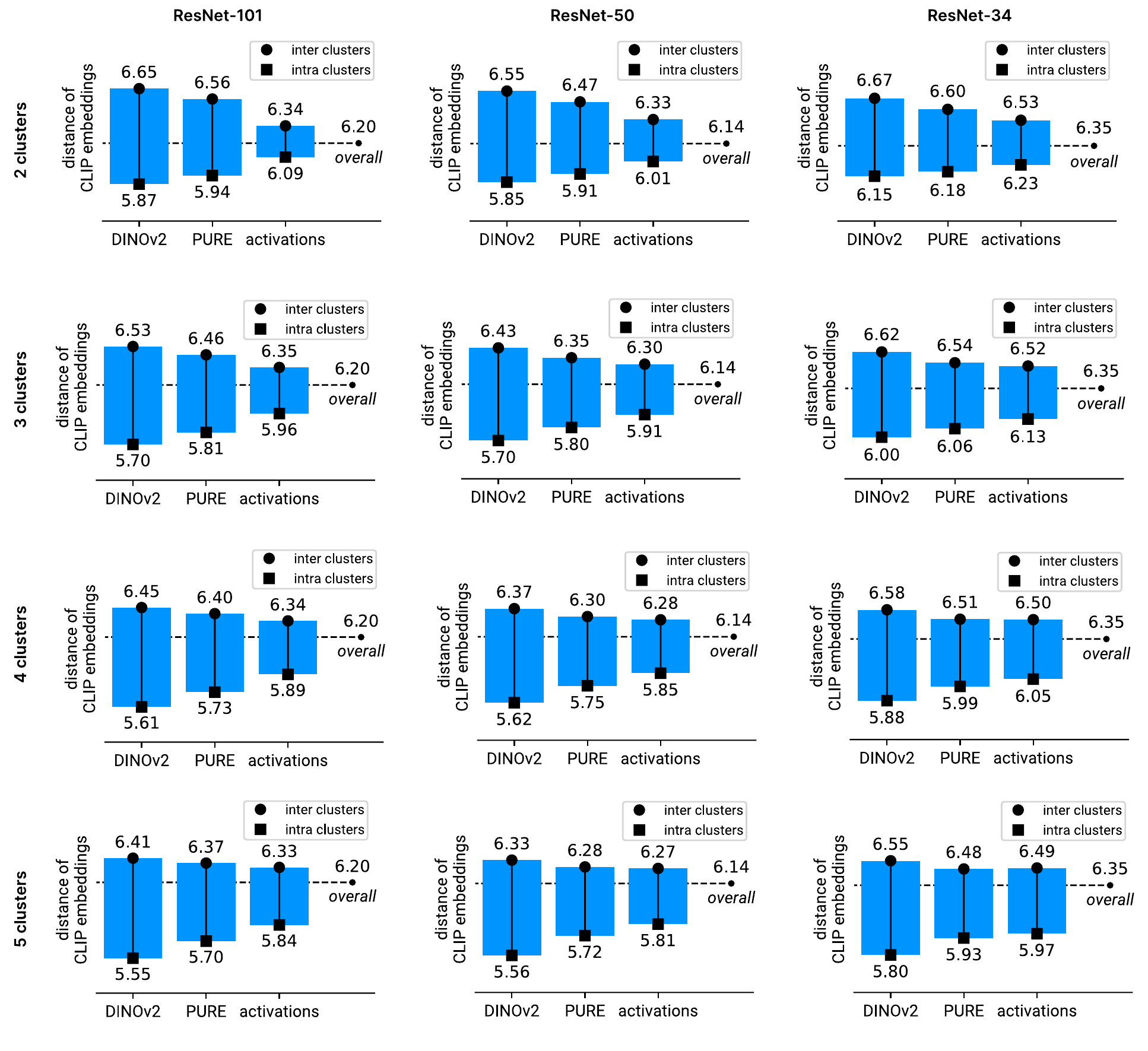}
        \caption{Results for measuring the degree of monosemanticity of clustered feature visualizations using CLIP embedding distances, as discussed in \cref{sec:evaluation:monosemanticity} for all ResNet architectures and different number of clusters (virtual neurons).}
        \label{fig:appendix:evaluation:monosemanticity}
    \end{figure*}

    In \cref{sec:evaluation:monosemanticity},
    we evaluate the effectiveness of disentanglement via CLIP.
    Specifically,
    we create $k$ new virtual neurons for each neuron by clustering the 100 most activating reference samples of a neuron into $k$ clusters.
    Then,
    the reference samples of each cluster are evaluated using CLIP,
    where ideally, the CLIP embedding distance decreases inside clusters and increases across clusters, indicating well (visually) separated clusters.
    In \cref{fig:appendix:evaluation:monosemanticity}, we present additional results for the main manuscript (where $k=2$ and ResNet-101 model results are shown)
    for ResNet-34 and ResNet-50 for $k\in\{2,3,4,5\}$.
    For all experiments,
    \gls{ours} leads to better cluster seperation than activation-based clustering.
    Notably,
    the higher $k$ is, the lower intra-cluster distances are in general, indicating visually more monosemantic feature visualizations per virtual neuron.
    However,
    inter-cluster distance often decreases when $k$ is increased,
    which indicates that most neurons are rather monosemantic, as also discussed in \cref{sec:appendix:distribution}.
    
\begin{figure}[t] 
        \centering 
        \includegraphics[width=1.0\linewidth]{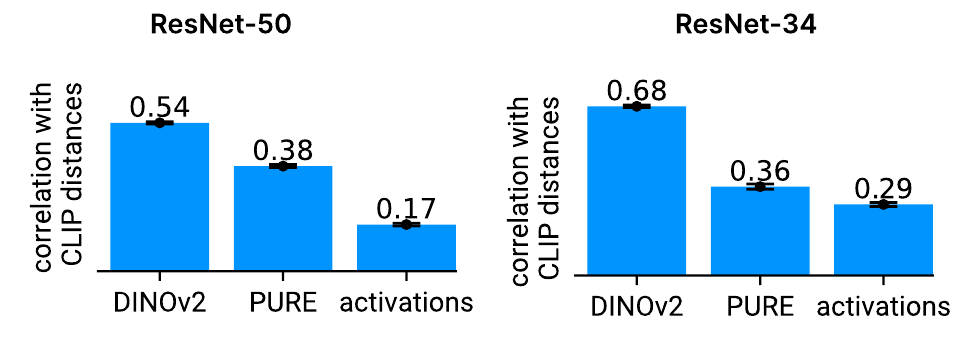}
        \caption{Correlation between feature visualization distances of CLIP to other methods for ResNet-50 (\emph{left}) and ResNet-34 (\emph{right}), which extends the results given in \cref{fig:quantitative}.}
        \label{fig:appendix:evaluation:correlation}
    \end{figure}

    A clustering that is aligned with CLIP results from similar distances between feature visualization pairs according to the respective methods.
    Concretely,
    CLIP and DINOv2 embeddings $\mathbf{e}_i^\text{CLIP}$ (and $\mathbf{e}_i^\text{DINOv2}$) are computed using feature visualizations (cropped reference samples) as the input for reference sample $i$ \wrt to a neuron $p$ in layer $L$.
    For \gls{ours} distances are computed on lower-level layer attributions $\mathbf{R}^{L-1}$ as given by \cref{eq:node_relevance_simple} when explaining neuron $p$.
    Further,
    for activation-based distances,
    activations $\mathbf{A}^{L}$ in layer $L$ are used, as proposed by \cite{O'Mahony_2023_CVPR}.
    The more aligned to CLIP,
    the higher the correlation of distances between the methods is.
    We provide additional results for \cref{sec:evaluation:monosemanticity} with results for ResNet-50 and ResNet-34 in \cref{fig:appendix:evaluation:correlation},
    for which \gls{ours} also shows higher correlations than activation.
    The correlation analysis involves examining the pairwise distances for the top-50 most activating reference samples across all neurons in the penultimate layer.
    The standard error of mean is computed by partitioning distances in 30 subsets (over which the mean is computed).
    

    \subsubsection{When Disentanglement Diverges from CLIP}

    In the following,
    we present examples, when activation-based or \gls{ours} attribution-based clustering of reference samples diverges from how CLIP embeddings cluster feature visualizations for the ResNet-50.
    
    \paragraph{Activation}
    We observe unfaithful clustering with activations
    when a significant portion of feature visualizations is of a single class, leading to high activation similarities (due to similar features present in the reference samples).
    The reference samples of the same class dominate in clustering, leading to all samples from different classes being clustered in another cluster.
    This is given, \eg, for neurons \texttt{\#143} (class ``custard apple''), \texttt{\#385} (class ``shoji''), \texttt{\#614} (class ``lacewing''), or \texttt{\#1055} (class ``buckeye'').
    Further, all reference samples can have small activation similarity if, \eg, most are from different classes, which is the case for neuron \texttt{\#1147}, leading to noisy clustering.
    
            \begin{figure*}[t]
        \centering 
        \includegraphics[width=0.95\linewidth]{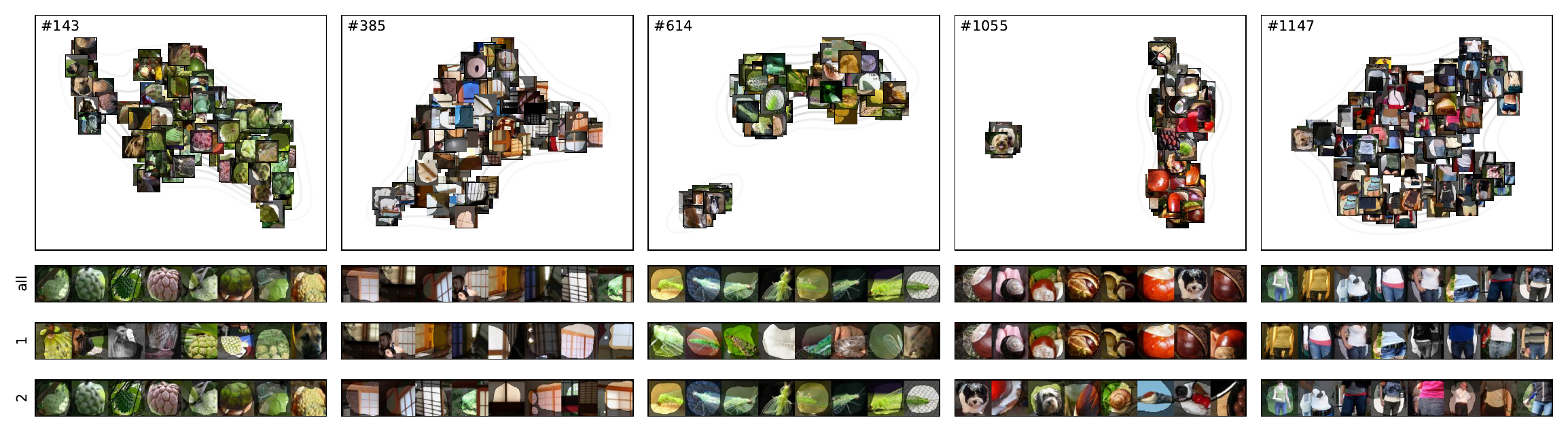}
        \includegraphics[width=0.95\linewidth]{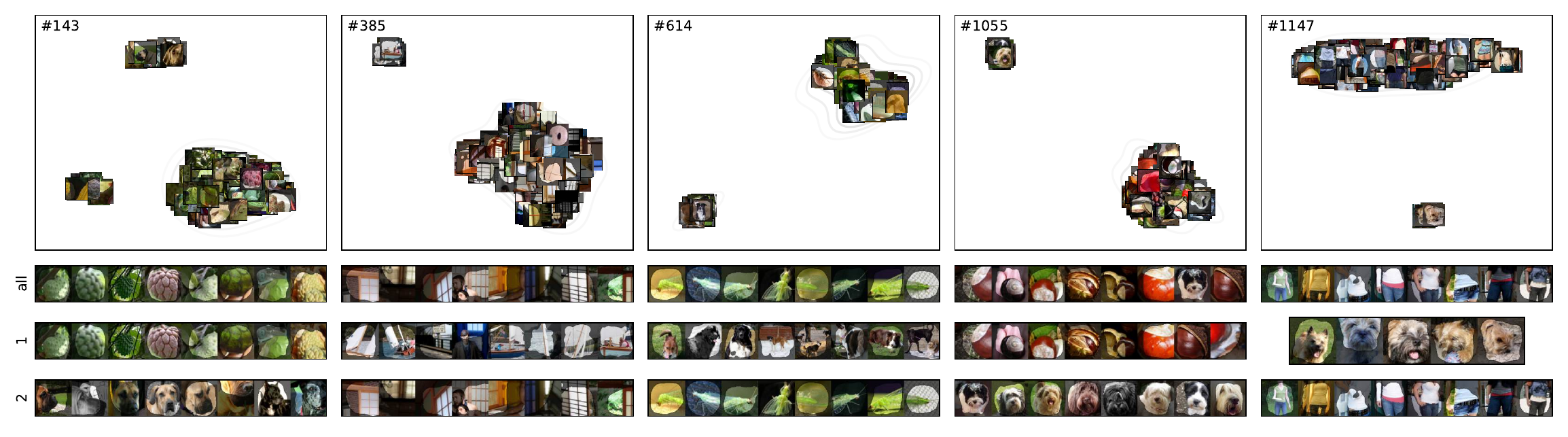}
        \caption{Examples for diverging disentanglement using activations compared to CLIP embeddings for the ResNet-50 model. We show UMAP embeddings and the corresponding feature visualizations before (``all'') and after (cluster ``1'' and ``2'') disentanglement using activations (\emph{top}) and CLIP embeddings (\emph{bottom})}
        \label{fig:appendix:evaluation:failure_act}
    \end{figure*}
\begin{figure*}[t]
        \centering 
        \includegraphics[width=0.95\linewidth]{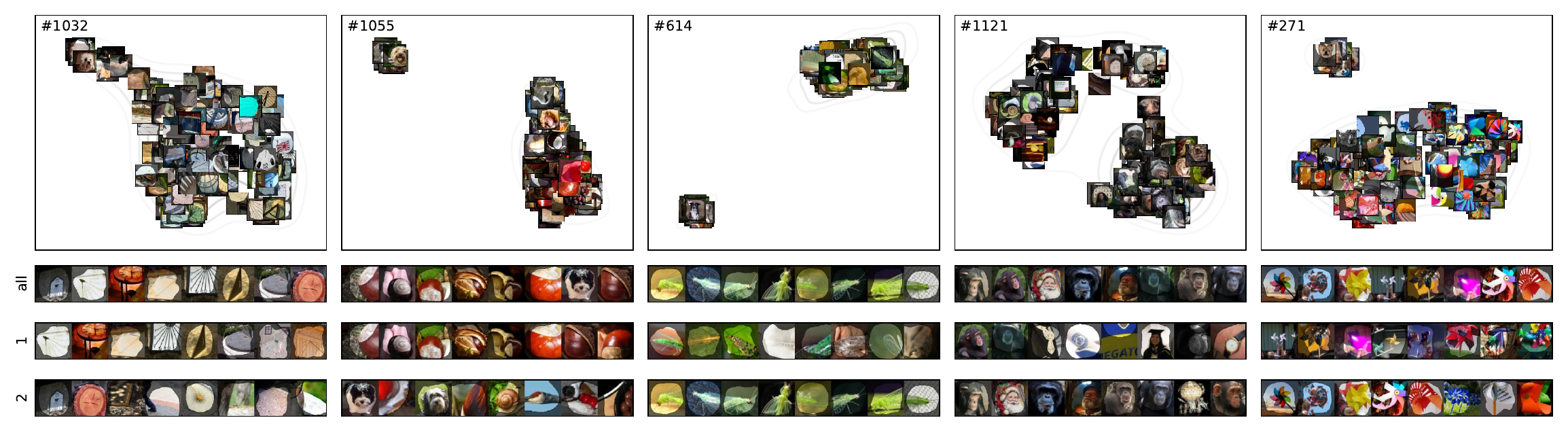}
        \includegraphics[width=0.95\linewidth]{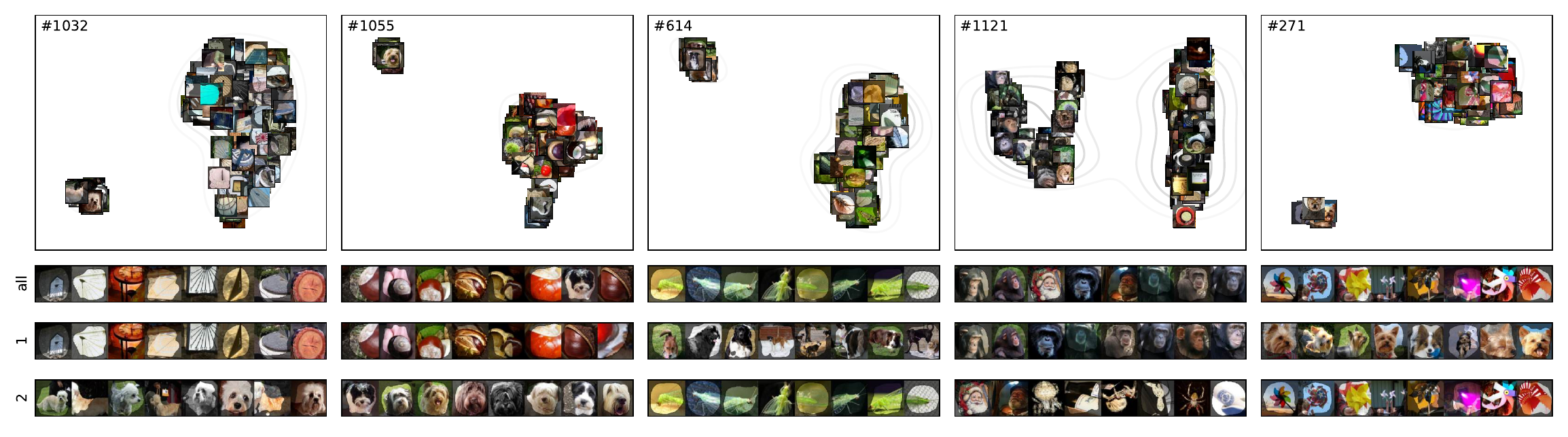}
        \caption{Examples for diverging disentanglement using \gls{ours} attributions compared to CLIP embeddings for the ResNet-50 model. We show UMAP embeddings and the corresponding feature visualizations before (``all'') and after (cluster ``1'' and ``2'') disentanglement using attributions (\emph{top}) and CLIP embeddings (\emph{bottom})}
        \label{fig:appendix:evaluation:failure_pure}
    \end{figure*}

    \paragraph{PURE}
    Similarly, as for activation,
    attributions can result in different clustering (compared to CLIP), \eg, for neurons \texttt{\#614} or \texttt{\#1055} as shown in \cref{fig:appendix:evaluation:failure_pure},
    where very small distances between reference samples from the same class  (``lacewing'' and ``buckeye'', respectively) lead to unaligned clustering.
    For other neurons, semantics can be more abstract and difficult to understand, such as \texttt{\#1032} (radially outspreading lines) or \texttt{\#1121}. 
    In these cases, CLIP seems to result in clustering reference samples corresponding to same depicted object classes (\eg, dogs or monkeys, respectively).
    This is also the case for neuron \texttt{\#271}, where the semantics seems to correspond to triangular shapes that can be found for toy windmills or dog ears.
    Here,
    CLIP separates dogs from windmills, even though they correspond to the same semantics,
    indicating also a disadvantage of using CLIP for evaluation.
    Especially for abstract concepts,
    attributions can result in low distances, whereas according to CLIP, bigger distances exist. 
    This also raises the question whether the ultimate goal of disentanglement is to separate clusters according to \emph{visual} difference or \emph{semantic} difference (as seen by the model).

\end{document}